\documentclass[letterpaper]{article} 
\usepackage{aaai2026}  
\usepackage{times}  
\usepackage{helvet}  
\usepackage{courier}  
\usepackage[hyphens]{url}  
\usepackage{graphicx} 
\usepackage{natbib}  
\usepackage{caption} 
\usepackage{caption}
\usepackage{graphicx}
\usepackage{xspace}
\usepackage{amsmath}
\usepackage{algorithm}
\usepackage{algorithmic}
\usepackage{booktabs}
\usepackage{enumitem}
\usepackage{amssymb}
\usepackage[many]{tcolorbox}
\definecolor{mypink}{rgb}{0.858, 0.188, 0.478}
\newcommand{\name}[0]{ToolACE-R\xspace}

\usepackage{newfloat}
\usepackage{listings}
\DeclareCaptionStyle{ruled}{labelfont=normalfont,labelsep=colon,strut=off} 
\floatstyle{ruled}
\newfloat{listing}{tb}{lst}{}
\floatname{listing}{Listing}
\title{ToolACE-R: Model-aware Iterative Training and Adaptive Refinement for \\ Tool Learning}
\author{
    Xingshan Zeng{$^{\rm 1}$}, Weiwen Liu{$^{\rm 2}$}\thanks{Corresponding Author.}, Xu Huang{$^{\rm 3}$},  Zezhong Wang{$^{\rm 1}$}, Lingzhi Wang{$^{\rm 4}$}, Liangyou Li{$^{\rm 1}$},\\ Yasheng Wang{$^{\rm 1}$}, Lifeng Shang{$^{\rm 1}$}, Xin Jiang{$^{\rm 1}$},  Ruiming Tang{$^{\rm 1}$}, Qun Liu{$^{\rm 1}$}
}
\affiliations{
    \textsuperscript{\rm 1}Huawei Technologies Co., Ltd\\
    \textsuperscript{\rm 2}Shanghai Jiao Tong University\\
    \textsuperscript{\rm 3}University of Science and Technology of China\\
    \textsuperscript{\rm 4}Harbin Institute of Technology, Shenzhen\\
zeng.xingshan@huawei.com,wwliu@sjtu.edu.cn
}

\usepackage{bibentry}

\begin{document}

\maketitle

\begin{abstract}
Tool learning, which allows Large Language Models (LLMs) to leverage external tools for solving complex user tasks, has emerged as a promising avenue for extending model capabilities. However, existing approaches primarily focus on data synthesis for fine-tuning LLMs to invoke tools effectively, largely ignoring how to fully stimulate the potential of the model.
In this paper, we propose \name, a novel framework that includes both model-aware iterative training and adaptive refinement for tool learning. \name features a model-aware iterative training procedure that progressively adjust training samples based on the model’s evolving capabilities to maximize its potential. Additionally, it incorporates self-refinement training corpus which emphasizes LLM's ability to iteratively refine their tool calls, optimizing performance without requiring external feedback. 
Furthermore, we introduce adaptive self-refinement for efficient test-time scaling, where the trained model can autonomously determine when to stop the process based on iterative self-refinement.
We conduct extensive experiments across several benchmark datasets, showing that \name achieves competitive performance compared to advanced LLMs. The performance can be further improved efficiently through adaptive self-refinement. These results highlight the effectiveness and generalizability of \name, offering a promising direction for more efficient and scalable tool learning.
\end{abstract}


\section{Introduction}
Tool learning, enabling Large Language Models (LLMs) to leverage external tools to address complex user requirements, has gained increasing attention. With tool integration, LLMs can access up-to-date information, perform intricate computations, and utilize third-party services, significantly expanding their capabilities beyond simple natural language communication with humans~\cite{qu_tool_2024}.
While tool invocation requires the LLMs to demonstrate strong understanding, reasoning, and instruction-following skills, customized fine-tuning is currently the dominant approach for enabling models to call external tools~\cite{liu2024toolacewinningpointsllm,liu2024apigen,patil2023gorilla,qin2023toolllm}. 

Due to the limited availability of high-quality data, existing research has primarily focused on developing effective and efficient methods for data synthesis using advanced models~\cite{abdelaziz2024granite,liu2024toolacewinningpointsllm,liu2024apigen,wang2024toolflow,prabhakar2025apigen}. However, data synthesized by advanced models from different sources can lead to compatibility issues. Specifically, when synthesized samples exceed the model's current knowledge, they may undermine the model's performance or lead to hallucinations~\cite{kang2024unfamiliar,ren_learning_2024}.
As a result, determining which training data samples are \textit{appropriate} for a given model remains an unresolved challenge.

Additionally, the potential for \textit{maximizing a model's intrinsic capabilities} remains underexplored in the field of tool learning. 
On one hand, prior work has seldom investigated how to fully exploit existing training data through techniques such as data augmentation. 
On the other hand, scaling test-time computation -- despite its demonstrated effectiveness in enhancing LLM reasoning~\cite{brown2024large,snell_scaling_2024} -- has received limited attention in tool learning scenarios.
A promising approach involves iteratively refining the model’s outputs and selecting the majority answer~\cite{snell_scaling_2024}. 
However, user queries vary widely in complexity, ranging from simple to highly intricate. 
Scaling test-time computation indiscriminately is inefficient, as models can often correctly answer straightforward questions without the need for additional refinement.
Therefore, an adaptive strategy for scaling test-time computation is needed to dynamically determine the appropriate level of computational effort required for each query.

To address the challenges outlined above, we propose \name, a novel pipeline for stimulating model potential in tool learning, encompassing both training and inference procedures.
Specifically, we propose a model-aware iterative training framework that enables LLMs to learn tool invocation in alignment with their evolving capabilities. This is achieved through a novel model-aware difficulty metric, guiding the training process in an iterative manner. Additionally, a simple yet effective self-refinement data incorporation strategy is employed as a form of data augmentation. Furthermore, by preserving refinement samples that include identical cases, LLMs are encouraged to learn appropriate stopping criteria during the iterative process. Building upon this training procedure, we integrate adaptive self-refinement into iterative inference, allowing models to autonomously determine when to halt refinement, thereby improving inference-time efficiency.

We have conducted extensive experiments on several representative tool-calling benchmarks, such as the Berkeley Function Call Leaderboard (BFCL)~\cite{berkeley-function-calling-leaderboard} and API-Bank~\cite{li_api-bank_2023}, to assess the effectiveness and efficiency of \name. Experimental results demonstrate that \name achieves competitive performance compared to advanced API-based models such as GPT-4o, and can be further enhanced through adaptive self-refinement. Additional analyses highlight the contributions of the proposed modules and examine the generalizability of our method across different model backbones and sizes.

The contributions of this work are summarized as follows:
\begin{itemize}[left=0pt]
    \item We propose a model-aware iterative training procedure, augmented with self-refinement corpus construction, to maximize the potential of LLMs for tool learning. 
    \item Building upon the training process, we integrate adaptive self-refinement into iterative inference, enabling models to autonomously decide when to halt the refinement process and allowing for more efficient use of computational resources when scaling test-time compute.
    \item We conduct extensive experiments on several representative tool calling benchmarks, revealing effectiveness and efficiency of our method.
\end{itemize}

\section{Related Work}
\paragraph{Tool Learning.}
Tool learning methods generally fall into two categories, tuning-free and tuning-based approaches. Tuning-free methods employ prompting techniques 
without additional training \cite{mialon2023augmented,hsieh2023tool,ruan2023tptu}. A notable example method is ReAct \cite{yaoreact}, which can alternate between reasoning and action for complex tasks. Recent work~\cite{qu_exploration_nodate} improves tool understanding by iteratively refining documentation.  
Tuning-based methods, which directly enhance tool calling ability via specialized fine-tuning, have recently gained much more attention~\cite{qin2023toolllm,schick2023toolformer,patil2023gorilla,tang2023toolalpaca,abdelaziz2024granite}.
A large amount of them focus on data synthesis for improved generation procedures \cite{liu2024toolacewinningpointsllm,liu2024apigen,wang2024toolflow}, often overlooking the model's inherent potential. 
\citet{zeng2025itool} also apply iterative training but based on DPO~\cite{rafailov_direct_2023} which introduces more complexity.
While some others integrate tool feedback for refinement~\cite{du_anytool_2024,wang_llms_2024}, our approach is more streamlined, directly refining outputs in an iterative manner without external or any textual feedback to maximize model efficiency.

\paragraph{Data Selection.}
Selecting high-quality training samples is essential for fine-tuning LLMs~\cite{albalak_survey_2024}.
A small set of high-quality data can effectively harness the model's potential, rather than relying on large quantities of data~\cite{zhou_lima_2023,liu_what_2023}. While earlier works primarily emphasize general data quality aspects, such as diversity and complexity, recent studies advocate for model-specific data selection~\cite{du2023mods,li-etal-2024-quantity}. This kind of approach is based on the observation that data distributions that significantly deviate from the base model's are challenging for the model to learn from and may even degrade the performance~\cite{ren_learning_2024}. Our method aligns with this perspective, defining a new metric to identify and select suitable training samples based on the model to be trained.

\paragraph{Self Refinement.}
Previous research has demonstrated that LLMs can refine their own generations through either self-feedback~\cite{madaan_self-refine_2023, weng_large_2023} or external feedback~\cite{qu_recursive_2024, xu_interactive_2024}. Nevertheless, it remains challenging for LLMs to assess the correctness of their refined output autonomously~\cite{huanglarge2024}, where scaling test-time compute with iterative refinement still often depends on post-processing techniques, such as majority voting~\cite{snell_scaling_2024}. In this work, we propose a simple yet effective method that enables LLMs to adaptively self-refine their outputs, improving scaling efficiency.

\begin{figure*}[t]
    \centering
\includegraphics[width=0.95\linewidth]{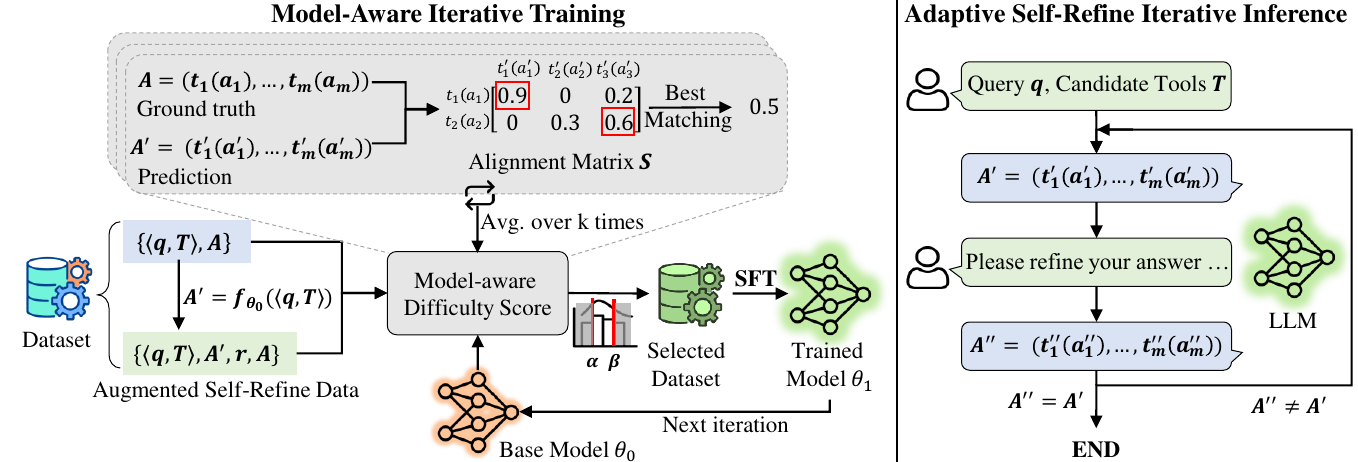}
    \caption{The overview of \name, including two components, for training and inference, respectively.}
    \label{fig:method}
\end{figure*}
\section{Methodology}
\subsection{Problem Formulation}
Given a user query \( q \) and a set of candidate tools \( T = \{t_1, t_2, \dots, t_n\} \), the objective of tool learning is to generate the correct tool invocations. This includes selecting the most appropriate tools and extracting the suitable parameter values. Specifically, the goal is to produce:
\begin{equation}
A = [t_1(a_1), \dots, t_m(a_m)] = f_\theta(\langle q, T \rangle)
\end{equation}
where \( t_j \) and \( a_j \) represent the \( j \)-th invoked tool and its corresponding argument, respectively, with \( 1 \leq j \leq m \) and \( m \) being the total number of tool invocations needed. The function \( f_\theta(\cdot) \) denotes the generation process of the LLM with parameters \( \theta \).
For each $a_j$, it might include several parameters and the corresponding values, denoted as $a_j = [p_1:v_1,...,p_i:v_i,...]$, where $p_i$ is the parameter name, and $v_i$ is the corresponding value.

The training samples for tool learning typically leverage $q$ and $T$ as context, and $A$ as output (the ground-truth tool invocation), denoted as $\{\langle q, T\rangle, A\}$.

\name is trained with a model-aware iterative training pipeline and can iteratively self-refine during inference in an adaptive manner. Figure~\ref{fig:method} shows the overall procedure.

\subsection{Model-Aware Iterative Training}

We begin model training with an instruction-tuned base model to ensure it possesses fundamental instruction-following capabilities required by our model-aware module. For the training process, we first collect a set of off-the-shelf samples and augment them via self-refinement. Rather than using these samples directly, we apply a selection criterion grounded in our defined notion of model-aware difficulty to curate the data. This curated dataset is then used to train the model and is iteratively re-curated based on new trained model throughout the training process.

\paragraph{Model-Aware Difficulty.}
We introduce a \textit{model-aware difficulty} metric that quantifies the inherent challenge a specific training sample poses to a given model. Rather than relying solely on static heuristics or dataset-level statistics, our definition measures how well the model can approximate the desired tool invocation behavior before training, thereby capturing a performance-aware notion of difficulty.

Let a training sample be denoted as $\{\langle q, T \rangle, A\}$, where $A = [t_1(a_1), \dots, t_m(a_m)]$ is the reference tool invocation sequence. Given a base model $\theta_0$, we obtain the model’s prediction as $A^{\prime} = [t^{\prime}_1(a^{\prime}_1), \dots, t^{\prime}_{m^{\prime}}(a^{\prime}_{m^{\prime}})] = f_{\theta_0}(\langle q, T \rangle)$.
To assess the similarity between the predicted and reference sequences, we adopt Jaccard-style overlap to define a \textit{parameter-level alignment score} between individual tool calls $t_i(a_i)$ and $t^{\prime}_j(a^{\prime}_j)$ as follows:
\begin{equation}
S_{ij} = \begin{cases}
\frac{|a_i \cap a^{\prime}_j|}{|a_i \cup a^{\prime}_j|} & \text{if } t_i = t^{\prime}_j \\
0 & \text{otherwise } 
\end{cases}
\label{eq:param-match}
\end{equation}
Here, $|a_i \cap a^{\prime}_j|$ denotes the number of matching arguments between $a_i$ and $a^{\prime}_j$, i.e., parameters with both identical names and values. $|a_i \cup a^{\prime}_j|$ represents the total count of distinct arguments across both calls. Parameters with the same name but differing values are counted separately to penalize incorrect assignments.

We aggregate all pairwise scores into a similarity matrix $S \in \mathbb{R}^{m \times m^{\prime}}$. To compute the best structural alignment between $A$ and $A^{\prime}$, we frame the task as a bipartite matching problem over the index sets $\mathcal{I} = {1, \dots, m}$ and $\mathcal{J} = {1, \dots, m'}$, and apply the Hungarian algorithm~\cite{kuhn1955hungarian} to identify the optimal matching $\mathcal{M}$ that maximizes the total alignment score:
\begin{equation}
\mathcal{M} = {\arg\max}_{\mathcal{M} \in \text{Matchings}(\mathcal{I}, \mathcal{J})} \sum_{(i,j)\in \mathcal{M}} S_{ij}
\end{equation}
where $\text{Matchings}(\mathcal{I}, \mathcal{J})$ denotes the set of all one-to-one matchings between $\mathcal{I}$ and $\mathcal{J}$, i.e., subsets of $\mathcal{I} \times \mathcal{J}$ in which each element in $\mathcal{I}$ and $\mathcal{J}$ appears in at most one pair.
Using this optimal alignment, we define the \textit{overlap score} between $A$ and $A^{\prime}$ as:
\begin{equation}
O(A, A^{\prime}) = \frac{\sum_{(i,j)\in \mathcal{M}} S_{ij}}{\max(m, m^{\prime})}
\end{equation}
This score reflects the degree to which the model reproduces the reference tool call, normalized by the longer of the two sequences to account for missing or superfluous tool calls.

To incorporate model uncertainty and account for variance in its outputs, we evaluate the model over $k$ independent attempts, resulting in outputs ${A^{\prime (1)}, \dots, A^{\prime (k)}}$. The \textit{model-aware difficulty} of a sample is then defined as:
\begin{equation}
D(\langle q, T\rangle, A) = 1 - \frac{1}{k} \sum_{l=1}^{k} O(A, A^{\prime (l)})
\label{eq:difficulty}
\end{equation}
This difficulty score $D \in [0, 1]$ quantifies the average divergence between the model’s output and the reference across multiple attempts. A lower score indicates that the model consistently produces accurate or nearly accurate tool calls, while a higher score suggests that the model struggles with the sample, failing to replicate even partial correctness. For instance, $D = 0$ implies perfect performance across all outputs, whereas $D = 1$ reflects complete failure in reproducing any reference calls.

This model-aware difficulty serves as a valuable signal for iterative training, enabling more effective model development in tool-use scenarios.

\paragraph{Data Augmentation with Self-Refinement.}
To augment the training corpus and establish the model's self-refinement capability, we additionally construct self-refinement data based on the original training samples. Specifically, we create a self-refinement sample as a multi-turn interaction, where the first turn is the initial response from the model, and the second turn includes a refinement prompt from the user followed by the refined answer (typically the ground-truth tool invocation $A$): $ \{ \langle q, T \rangle, A_1, r, A \} $. Here, the model to be trained will first produce its output tool invocation $A_1$. We then concatenate a refinement requirement prompt $r$ (a generalized sentence like ``Please refine your answer...'') and the final ground-truth answer $A$.

Notably, there may be samples where $A_1 = A$, meaning the refined answer is identical to the previous answer. We retain these samples to enable the model to learn that "no further changes are needed when the answer is perfect." This is a crucial aspect of establishing the model's adaptive, iterative self-refinement, and we will provide further details in later section.

Each self-refinement sample can also be estimated its difficulty using Eq.~\ref{eq:difficulty}, by replacing $\{\langle q, T\rangle\}$ with $\{\langle q, T\rangle, A_1, r\}$ as context.
In this way, all the training samples including the constructed self-refinement samples can be estimated and labeled with a difficulty score.

\paragraph{Model-Aware Iterative Data Selection and Training.}
We hypothesize that the most suitable training samples are those within the model's current capability, either too difficult or too simple samples are ineffective for model training.
To that end, we empirically set two thresholds $\alpha$ and $\beta$, where $\alpha < \beta$. For any training sample $\{C, A\}$ (where $C$ indicates the context, $\langle q, T\rangle$ or $\{\langle q, T\rangle, A_1, r\}$ for self-refinement data), if its estimated difficulty score $\alpha < D(C, A) < \beta$, we retain it as an effective training sample and add to the actual training set $\mathcal{D}_0$.

Given the selected training set $\mathcal{D}_0$, we apply supervised fine-tuning to the base model $\theta_0$ for one epoch, producing $\theta_1$. This constitutes the first iteration. In subsequent iterations, we treat the trained model as the new base model, re-construct the self-refinement samples, re-evaluate the difficulty, and re-select samples with the same criterion for fine-tuning the updated model. 

Ideally, with each iteration, the model's performance improves, leading to an update of data difficulty and so new training samples might be included matching model's current capability. The iteration will continue until the performance improvement is marginal, indicating that the model has saturated on this set of training samples. The resulting model will then be our final \name model.

\subsection{Adaptive Self-Refine Iterative Inference}
\label{sec:method:inf}
With model-aware iterative training, our \name model can directly generate tool invocations based on a user query and candidate tools, and refine its answer when necessary.

We propose an adaptive self-refine procedure that enables \name to iteratively refine its own output and autonomously determine when to halt the iteration. Details are shown in Figure~\ref{fig:method} (right hand side).
Specifically, given a user query $q$ and candidate tools $T$, \name first generates the tool invocation $A^{\prime}$. In each subsequent iteration, \name leverages the answer from the previous iteration as context and refines it. As mentioned earlier, \name has learned that when the answer is sufficiently accurate, it can produce the same answer in subsequent refinement. Therefore, the iteration continues until the answers from two consecutive iterations are identical. We refer to this procedure as adaptive self-refinement, where the model itself determines when to stop the iterations. To prevent infinite iteration, we set a maximum iteration limit $n$.
Through adaptive self-refinement, \name is able to improve its performance, particularly when facing difficult cases.

\section{Experiments}

\subsection{Experimental Settings}
\paragraph{Datasets and Models.}
We use a subset of ToolACE~\cite{liu2024toolacewinningpointsllm} training data as our training data, where we only retain the samples with single-turn tool calling for simplicity. 
We select LLaMA3.1-8B-Instruct~\cite{llama3modelcard} as base model in our main experiments. We also conduct experiments on Qwen2.5-Instruct-series~\cite{yang2024qwen2} (0.5B, 1.5B, 3B and 7B) and Mistral-7B-Instruct-v0.2~\cite{jiang2023mistral}, to validate the generalizability of \name.
We compare with the state-of-the-art API-based models, including GPT-series
and Gemini-2.5-Pro,
as well as open-source models like DeepSeek-V3~\cite{liu2024deepseek}, Llama3.1-8B/70B-Instruct~\cite{llama3modelcard} and Qwen2.5-7B-Instruct~\cite{yang2024qwen2}. We also compare with fine-tuned tool calling models like ToolACE-8B~\cite{liu2024toolacewinningpointsllm}, Hammer2.1-7B~\cite{lin2024hammer} and xLAM-2-8b-fc-r~\cite{zhang_xlam_2024,prabhakar2025apigen}.

\paragraph{Benchmarks and Evaluation.}
We conduct experiments on several representative benchmarks, including the Berkeley Function Call Leaderboard (BFCL)~\cite{berkeley-function-calling-leaderboard}, ACEBench~\cite{chen2025acebench}, API-Bank~\cite{li_api-bank_2023} and ToolAlpaca~\cite{tang2023toolalpaca}.
To minimize external interference and simplifying data preparation, we mainly evaluate performance on single-turn tool calling queries. For BFCL, we evaluate the Non-live and Live subsets, which correspond to synthetic test cases and real-world scenarios, respectively. Each subset includes four categories: Simple, Multiple, Parallel, and Multiple Parallel.
For ACEBench, we evaluate only the English normal category, excluding multi-turn cases. This includes the test samples in atom category and single-turn category. For API-Bank and ToolAlpaca, we focus on the first step tool calling samples, disregarding further tool or retrieval feedback. All performance is reported in terms of accuracy, where only cases with entirely correct tool calling are counted as correct.

\begin{table*}[thb]
\small
\setlength{\tabcolsep}{4.0pt}
\renewcommand{\arraystretch}{1.0}
\centering
\begin{tabular}{@{}lccccccccccc@{}}
\toprule
& \multicolumn{4}{c}{\textbf{Non-Live}}          & \multicolumn{4}{c}{\textbf{Live}}              & \multicolumn{3}{c}{\textbf{Overall}} \\ \cmidrule(lr){2-5}\cmidrule(lr){6-9}\cmidrule(lr){10-12}
\textbf{Models} &
  \multicolumn{1}{c}{\textit{Simple}} &
  \multicolumn{1}{c}{\textit{Multiple}} &
  \multicolumn{1}{c}{\textit{Parallel}} &
  \multicolumn{1}{c}{\begin{tabular}[c]{@{}c@{}}\textit{Multiple}\\ \textit{Parallel}\end{tabular}} &
  \multicolumn{1}{c}{\textit{Simple}} &
  \multicolumn{1}{c}{\textit{Multiple}} &
  \multicolumn{1}{c}{\textit{Parallel}} &
  \multicolumn{1}{c}{\begin{tabular}[c]{@{}c@{}}\textit{Multiple}\\ \textit{Parallel}\end{tabular}} &
  \multicolumn{1}{c}{\textit{Non-live}} &
  \multicolumn{1}{c}{\textit{Live}} &
  \multicolumn{1}{c}{\textit{Overall}} \\ 
  \midrule

\textbf{GPT-4o-2024-11-20 }                   & 77.17&	95.00&	93.50&	85.00 & \underline{84.50}&	79.30&	\underline{87.50}&	70.83& 87.67 & 80.24 & 83.96 \\

\textbf{GPT-4.1-2025-04-14    }          	&\textbf{80.50}&	94.00&	93.00&	87.50&\textbf{85.66}	&76.54	&\textbf{93.75}	&75.00&88.75&78.46&83.60\\

\textbf{GPT-4o-mini-2024-07-18   }          	&\underline{80.08}&	90.50&	89.50&	87.00 	&81.40&	76.73	&\textbf{93.75}	&\underline{79.17} &86.77 & 77.87 & 82.32\\

\textbf{Gemini-2.5-Pro-Exp-03-25 }             & 	67.50 &	91.00	 &83.50	 &73.00 & 79.46&	64.96&	81.25&	\underline{79.17}& 78.75 & 68.17 & 73.46\\

\midrule
\textbf{DeepSeek-V3} & 	78.67 &	95.50&	91.00	&\textbf{91.50} & 83.72	&\textbf{82.15}	&81.25	&62.50 & 89.17& \textbf{82.09} & \underline{85.63} \\ 
\textbf{Llama3.1-70B-Inst} & 77.92 &	\underline{96.00} &	\underline{94.50}	&\textbf{91.50} & 78.29	& 76.16	& \underline{87.50}	& 66.67 &  \underline{89.98} & 76.53 & 83.26 \\ 	
\textbf{Llama3.1-8B-Inst}  &	71.50&	93.50&	86.50&	86.00&	73.26&	68.95&	56.25&	50.00& 	84.37&	69.28 & 76.83 
 \\
\textbf{Qwen2.5-7B-Inst}&	70.50&	94.00&	90.50&	82.50&	74.81&	71.89&	62.50&	62.50 & 84.38 &	72.17 & 78.28
 \\
\textbf{ToolACE-8B (FC)} & 	76.67	&93.50&	90.50	&89.50&73.26	&76.73&	81.25	&70.83& 87.54 & 76.02 & 81.78\\
\textbf{Hammer2.1-7B (FC)} & 	78.08&	95.00&	93.50 &	88.00 &76.74	&77.40&	81.25	&70.83	& 88.65  & 77.20 & 82.92\\
\textbf{xLAM-2-8b-fc-r (FC) }                           	&	73.08	&93.50&	87.00	&84.00&74.81&	66.29&	56.25&	50.00&84.40&67.51&75.95 \\
\midrule
\textbf{\name (FC)} &	79.75	&\textbf{97.00}	&\textbf{95.50}&	\textbf{91.50}&	82.95	&\underline{81.10}	&\textbf{93.75}	&\textbf{87.50}& \textbf{90.94}&	\underline{81.72}&\textbf{86.33}
 \\
\bottomrule
\end{tabular}
\caption{\label{tab:bfcl-overall}Accuracy comparison on BFCL (Last updated on 2025-04-25).  The best results in each category are marked in \textbf{bold}. The second best results are \underline{underlined}. FC indicates the models are fine-tuned for function calling. }

\end{table*}

\paragraph{Implementation Details.}
We employ parameter-efficient fine-tuning method LoRA~\cite{hu2022lora} given resource constraints. All model modules are configured for LoRA fine-tuning, with a rank of 16 and an alpha value of 32. Training is performed with a global batch size of 64 and a learning rate of \(1 \times 10^{-4}\), following a cosine learning rate schedule with a warmup ratio of 0.1.  
For difficulty estimation, we set the generation temperature to 1.0 to encourage diversity and allow the model to explore a broader range of outputs. Model generates 8 times (i.e. $k=8$) for each sample to ensure robustness.
During data selection, we set $\alpha=0$ and $\beta=0.9$, which means that we discard samples that are extremely easy (difficulty score equals to $0$, i.e., the model always answer fully correctly) or too difficult (most attempts are total incorrect).
For the constructed refinement data, we apply a loss mask to the first assistant turn of the conversation, as it may involve an incorrect tool invocation, making it unsuitable for the model to learn from this turn.
During evaluation, we adopt greedy search to ensure stability. 
Unless otherwise specified, , the maximum iteration time for adaptive self-refinement is set to \(n = 5\) .

\subsection{Main Results}
\paragraph{Results on BFCL.}
Table~\ref{tab:bfcl-overall} presents the comparison results on BFCL, including detailed results for each category. 
The following key observations can be made:

\begin{itemize}[left=0pt]
    \item The performance gap between API-based and open-source models is minimal. Large open-source models like DeepSeek-V3 show comparable or even superior overall performance to GPT models and Gemini-2.5-Pro. 
    \item Specialized fine-tuned models significantly benefit from domain-specific fine-tuning, where smaller models (7/8B) are allowed to compete with larger general models like Llama3.1-70B-Inst.
    \item Our model, \name, which leverages our proposed model-aware iterative training procedure, achieves the best overall performance, surpassing both large API-based and open-source models.
    
\end{itemize}

\begin{figure}[t]
    \centering
\includegraphics[width=0.95\linewidth]{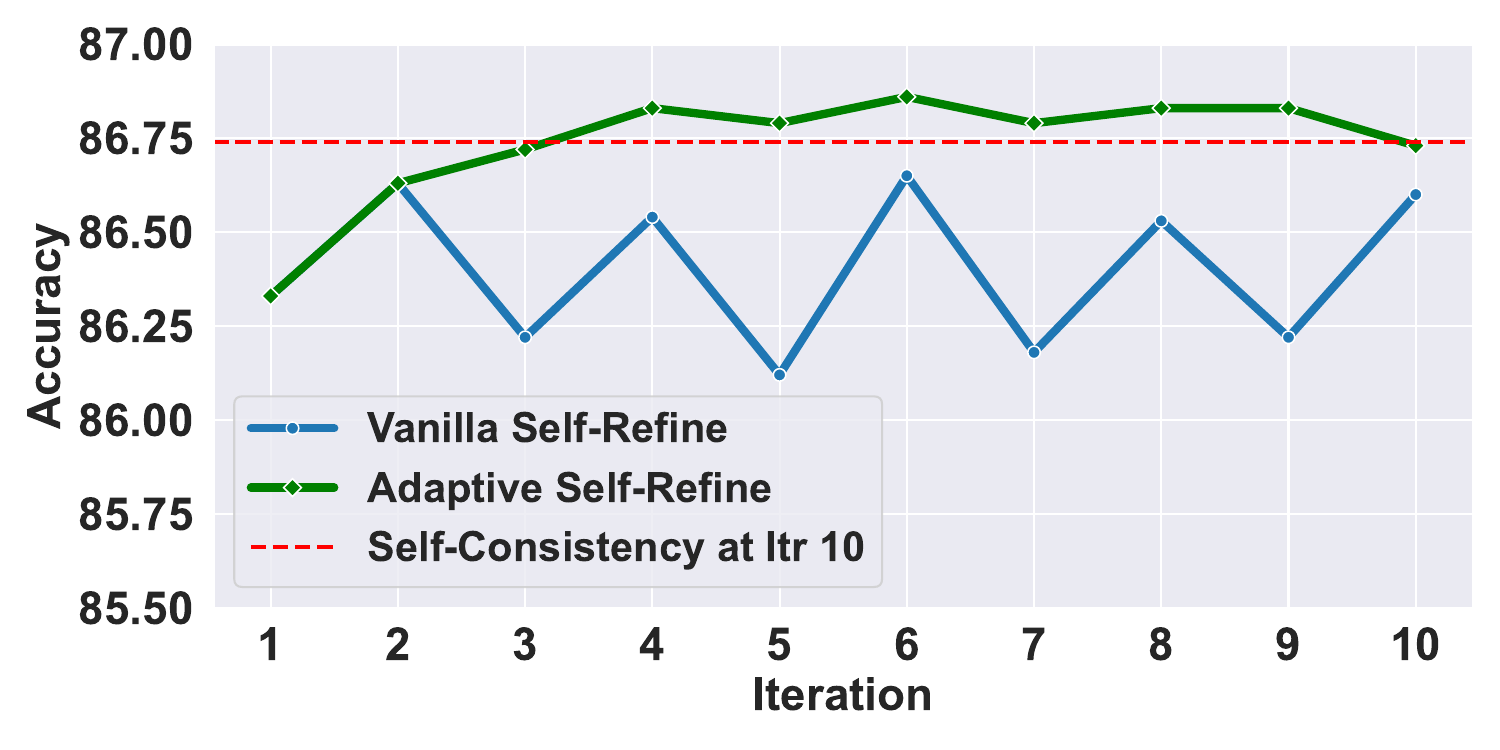}
    \caption{Performance of different iterative self-refine methods. ``Vanilla Self-Refine'' always picks the final answer at each iteration. ``Adaptive Self-Refine'' is our proposed method, while ``Self-Consistency'' selects the majority answer during the iteration.}
    \label{fig:inf-itr}
\end{figure}
To further assess the effectiveness of our adaptive self-refine method for iterative inference, we conducted experiments comparing with two other inference-time scaling approaches. The first, termed "Vanilla Self-Refine," always selects the last answer at each iteration as the final answer. The second method, "Self-Consistency at Itr n", chooses the majority answer from the answers collected during iterations 1 to n. The results of these inference methods on BFCL, 
are shown in Figure~\ref{fig:inf-itr}.

As observed, the performance of "Vanilla Self-Refine" fluctuates across iterations, indicating that the model does not always refine the answer correctly. In contrast, our "Adaptive Self-Refine" method demonstrates more stability across iterations, suggesting that adjacent identical answers serve as a reliable signal for stopping further iterations. 
Our statistics also show that the average iteration time of our method on BFCL is $2.4$ (when maximum iteration time $n=5$), indicating that most cases terminate early thus achieving consistent performance.
Our method achieves competitive performance compared to the strong baseline "Self-Consistency at Itr 10", further validating advantages.

\begin{table}[t]
\small
\centering
\setlength\tabcolsep{5.0pt}
\begin{tabular}{lccc}
\toprule
\multicolumn{1}{l}{\textbf{Models}} & \textbf{ACEBench} & \textbf{APIBank} & \textbf{ToolAlpaca} \\ 
\midrule
\multicolumn{1}{l}{\textbf{GPT-4o}} &\textbf{87.00}&\underline{77.16}&83.87\\
\multicolumn{1}{l}{\textbf{GPT-4o-mini}} &81.63&73.10&83.87\\ 
\multicolumn{1}{l}{\textbf{Llama3.1-8B-Inst}}&50.63&60.41&79.03\\
\midrule
\textbf{\name} &83.88&75.89&\underline{85.48} \\
 $\;\;$\textbf{+ Adaptive SR} &\underline{84.00}&\textbf{81.22}&\textbf{88.71}\\
\bottomrule
\end{tabular}
\caption{Accuracy comparison on ACEBench, API-Bank and ToolAlpaca. ``SR'' is short for ``Self-Refine''.}
\label{tab:other-overall}
\end{table}

\begin{table}[t]
\small
\centering
\setlength\tabcolsep{5.0pt}
\begin{tabular}{lccc}
\toprule
\multicolumn{1}{l}{\textbf{Models}} & \textbf{Non-live} & \textbf{Live} & \textbf{Overall} \\ 
\midrule
\textbf{\name + Adaptive SR} & \textbf{91.32} &\textbf{82.34}& \textbf{86.83}  \\
 \textbf{\name} & \underline{90.94} & \underline{81.72}& \underline{86.33} \\
 $\;\;$\textbf{- Data Selection} &88.96&81.13&85.04 \\
 $\;\;\;\;$\textbf{- SR Data} &88.75&80.09&84.42 \\
 \midrule
 \multicolumn{1}{l}{\textbf{Llama3.1-8B-Inst (base)} } & 84.37& 69.28 &76.83\\
\bottomrule
\end{tabular}
\caption{Ablation Study of our proposed modules on BFCL. ``SR'' is short for ``Self-Refine''. }
\label{tab:ablation}
\end{table}

\paragraph{Results on More Benchmarks.}
To provide a more comprehensive evaluation, we continue to conduct experiments on three other representative benchmarks ACEBench, API-Bank, and ToolAlpaca, focusing on single-turn tool-calling scenarios. The results are summarized in Table~\ref{tab:other-overall}.

On ACEBench, \name demonstrates an impressive absolute improvement of over 30\% compared to the base model, Llama3.1-8B-Inst, highlighting the effectiveness of our training pipeline. However, the benefit of Adaptive Self-Refine is marginal, likely due to the presence of particularly challenging cases where the model struggles to correctly refine its answers.

For both API-Bank and ToolAlpaca, \name achieves substantial improvements. Moreover, when combined with Adaptive Self-Refine, the performance sees further gains, even surpassing GPT-4o. These results validate the potential of Adaptive Self-Refine to progressively refine tool invocation answers based solely on the model's own capabilities, opening up new possibilities for scaling inference in tool learning scenarios.

\paragraph{Ablation Study.}
To evaluate the contribution of each proposed module, we conduct an ablation study by progressively removing them. The corresponding results on the BFCL dataset are presented in Table~\ref{tab:ablation}. As shown, the additional inference cost introduced by adaptive self-refinement leads to performance gains in both categories. Further removing model-aware data selection and the constructed self-refinement data both result in a clear performance drop, with the absence of model-aware data selection causing the most significant degradation. These findings demonstrate the effectiveness of our proposed modules, especially for the model-aware part which mainly comes from the novel definition of data sample difficulty.

\subsection{Effects of Model-Aware Iterative Training}
In this subsection, we explore several different choices when applying model-aware iterative training, to validate the effectiveness of our current proposal.

\paragraph{Choices of Selection Thresholds.}
In our main experiments, we set $\alpha=0$ and $\beta=0.9$ to discard samples that are either extremely simple or too difficult for model training. Here we experiment with more different choices of the two thresholds. We turn to another three kinds of choices, preserving simpler samples, preserving more difficult samples, or preserving samples in more medium difficulty. 
Table~\ref{tab:thresholds} displays the evaluation results on BFCL. 
As shown, setting $\alpha = -1$ and $\beta = 0.9$, which includes even extremely simple cases (i.e., $\alpha = 0$), leads to a slight performance drop while increasing training time due to the inclusion of a larger training corpus. In contrast, preserving only relatively difficult cases (by setting $\alpha = 0.1$ and $\beta = 2$) results in a notable performance degradation across both categories, with an overall decrease of approximately 1\%. This suggests that samples beyond the model’s current capability may hinder effective learning and limit performance improvements based on its existing potential. We further evaluate a stricter setting ($\alpha = 0.1, \beta = 0.8$), which discards more simple and difficult cases. This also results in performance drop, especially for Live category. The likely reason is that an excessive number of training samples are discarded (reducing the dataset to approximately half the size of our main setting) while retaining only a small portion of the harder examples.

\begin{table}[t]
\small
\centering
\setlength\tabcolsep{5.0pt}
\begin{tabular}{lccc}
\toprule
\multicolumn{1}{l}{\textbf{Thresholds}}  & \textbf{Non-live}  & \textbf{Live} & \textbf{Overall}  \\ 
\midrule
\multicolumn{1}{l}{\textbf{$\alpha=0,\beta=0.9$ (\name)}} &90.94&81.72&86.33\\
\multicolumn{1}{l}{\textbf{$\alpha=-1,\beta=0.9$}} &90.73&81.27&86.00 \\
\multicolumn{1}{l}{\textbf{$\alpha=0.1,\beta=2$}} &89.27&81.35&85.31\\
\multicolumn{1}{l}{\textbf{$\alpha=0.1,\beta=0.8$}} &90.83&80.38&85.61\\
\bottomrule
\end{tabular}
\caption{Performance on BFCL when applying different threshods during model-aware data selection. We always select samples that their difficulty score satisfy $\alpha < D < \beta$.  }
\label{tab:thresholds}
\end{table}

\begin{table}[t]
\small
\centering
\setlength\tabcolsep{5.0pt}
\begin{tabular}{lccc}
\toprule
\multicolumn{1}{l}{\textbf{Models}}  & \textbf{Non-live} & \textbf{Live}  & \textbf{Overall}  \\ 
\midrule
\multicolumn{1}{l}{\textbf{$k=8$ (\name)}}&90.94&81.72&86.33\\
\multicolumn{1}{l}{\textbf{$\;\;$+ Adaptive SR}} &91.32&82.34&86.83\\
\midrule
\multicolumn{1}{l}{\textbf{\textbf{$k=1$}}} &90.94&80.98&85.96 \\
\multicolumn{1}{l}{\textbf{$\;\;$+ Adaptive SR}}&90.71&82.01&86.36\\
\bottomrule
\end{tabular}
\caption{Performance on BFCL when setting $k$ value in Eq.~\ref{eq:difficulty} as $8$ and $1$. ``SR'' is short for ``Self-Refine''.}
\label{tab:k_value}
\end{table}

\paragraph{Choices of $k$.}
In Eq.~\ref{eq:difficulty}, we define the difficulty score by allowing the model to make $k$ attempts, thereby better exploring its potential and producing a more robust estimate. To evaluate the impact of different $k$ values, we compare our main setting ($k = 8$) with a reduced setting ($k = 1$). The corresponding results are shown in Table~\ref{tab:k_value}. The model trained with $k = 1$ exhibits a notable performance drop, particularly in the Live category, where difficult cases play a more critical role. Upon examining the final training set, we observe that approximately $10\%$ of the samples within the difficulty range $[0.8, 0.9]$ under the $k = 8$ setting are excluded when using $k = 1$. This provides evidence that a larger $k$ value yields a more stable difficulty estimation and facilitates the selection of more appropriate training samples.

\paragraph{Choices of Models.}
We further examine the ``model-aware'' part, where we use the base model to be trained to evaluate the difficulty of training samples.
We conduct an additional experiment. Instead of using data selected by the base model Llama3.1-8B-Inst, we train Llama3.1-8B-Inst with data selected by Qwen2.5-3B-Inst and Qwen2.5-7B-Inst, as well as with the intersection of all three models' selected data -- i.e., only the samples that all three models perform similar (within the same difficulty scope) are included in the dataset. The results of the trained models on BFCL, along with the amounts of corresponding training data (in the first iteration), are presented in Table~\ref{tab:model-aware}. As shown, performance drops when using data selected by other models, even when the data volume is larger. The intersection dataset, which contains less data, achieves comparable results. This observation underscores the benefits of model-aware data selection.

\begin{table}[t]
\small
\centering
\setlength\tabcolsep{5.0pt}
\begin{tabular}{lccc}
\toprule
\multicolumn{1}{l}{\textbf{Models}}  & \textbf{Live} & \textbf{Overall}   & \textbf{Amount}          \\ 
\midrule
\multicolumn{1}{l}{\textbf{\name}}&81.72&86.33&120K\\
\midrule
\multicolumn{1}{l}{\textbf{Select w/ Qwen2.5-3B}} &80.83&85.50&125K\\
\multicolumn{1}{l}{\textbf{Select w/ Qwen2.5-7B}} &81.27&85.97&134K\\
\multicolumn{1}{l}{\textbf{Intersection Data}} &81.37&86.08&108K\\
\bottomrule
\end{tabular}
\captionof{table}{Performance on BFCL when using different selected data to train Llama3.1-8B-Inst, along with the data amount. ``Qwen2.5-3/7B'' is short for Qwen2.5-3/7B-Inst.}
\label{tab:model-aware}
\end{table}

\begin{table}[t]
\small
\centering
\setlength\tabcolsep{5.0pt}
\begin{tabular}{lccc}
        \toprule
        \multicolumn{1}{l}{\textbf{Models}}         & \textbf{Non-live}            & \textbf{Live} & \textbf{Overall}    \\ 
        \midrule
        \multicolumn{1}{l}{\textbf{Qwen2.5-7B-Inst}} & 84.38 &72.17 &78.28\\
        \multicolumn{1}{l}{\textbf{$\;\;$+ Original Data}}&88.29&78.39&83.34\\
        \multicolumn{1}{l}{\textbf{$\;\;$+ \name}} &89.23&81.20&85.21\\
        \midrule
        \multicolumn{1}{l}{\textbf{Mistral-7B-Inst-v0.2}} & 55.25&55.07&55.16\\
        \multicolumn{1}{l}{\textbf{$\;\;$+ Original Data}} &84.06&70.17& 77.12\\
        \multicolumn{1}{l}{\textbf{$\;\;$+ \name}} & 87.98&80.01&84.00\\
        \bottomrule
        \end{tabular}
        \captionof{table}{Performance of using different models on BFCL. 
        }
\label{tab:diff-backbone}
\end{table}

\begin{figure}[t]
    \centering
\includegraphics[width=0.95\linewidth]{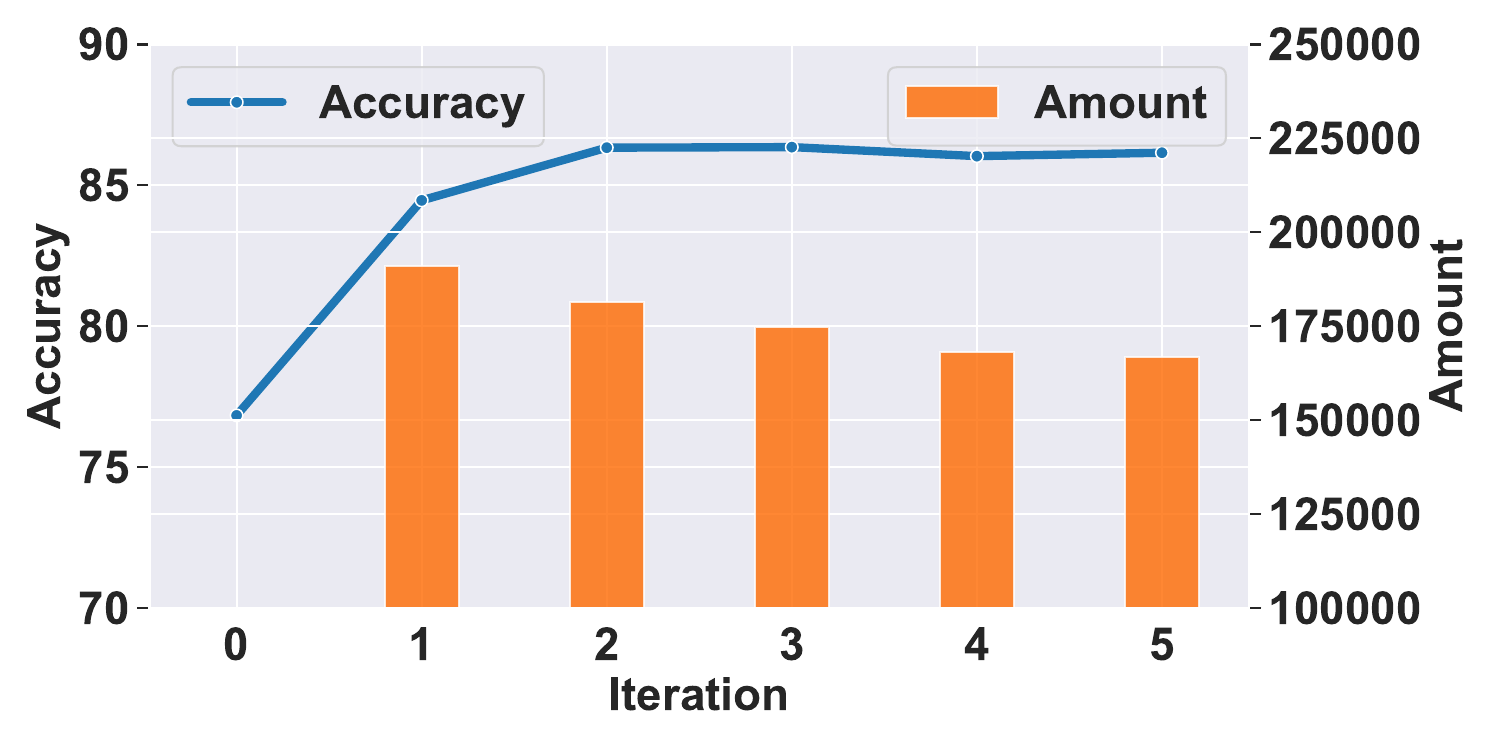}
    \caption{Performance and training amount of each iteration during training. Iteration 0 means the model before training.}
    \label{fig:train-itr}
\end{figure}

\paragraph{Iterative Training.}
To examine the training dynamics, we visualize the iterative training process by illustrating the trends in both accuracy and the amount of training data across iterations, as shown in Figure~\ref{fig:train-itr}. Iteration 0 corresponds to the base model, LLaMA3.1-8B-Inst. As depicted, the number of selected training samples decreases monotonically with each iteration, primarily because more samples are assigned a difficulty score of $0$ and consequently excluded from the training corpus.
Meanwhile, accuracy improves rapidly during the first two iterations, peaking at iterations 2 and 3. Beyond this point, performance plateaus, with no further gains observed in subsequent iterations. This pattern suggests that excessive training may lead to overfitting on the filtered dataset.
Therefore, to save the training cost, we recommend iterative training at most 3 times.

\subsection{Further analysis}
In this subsection, we examine the generalizability of our method by varying the base model and exploring the scaling effects across different model sizes.

\begin{figure}[t]
    \centering
\includegraphics[width=0.95\linewidth]{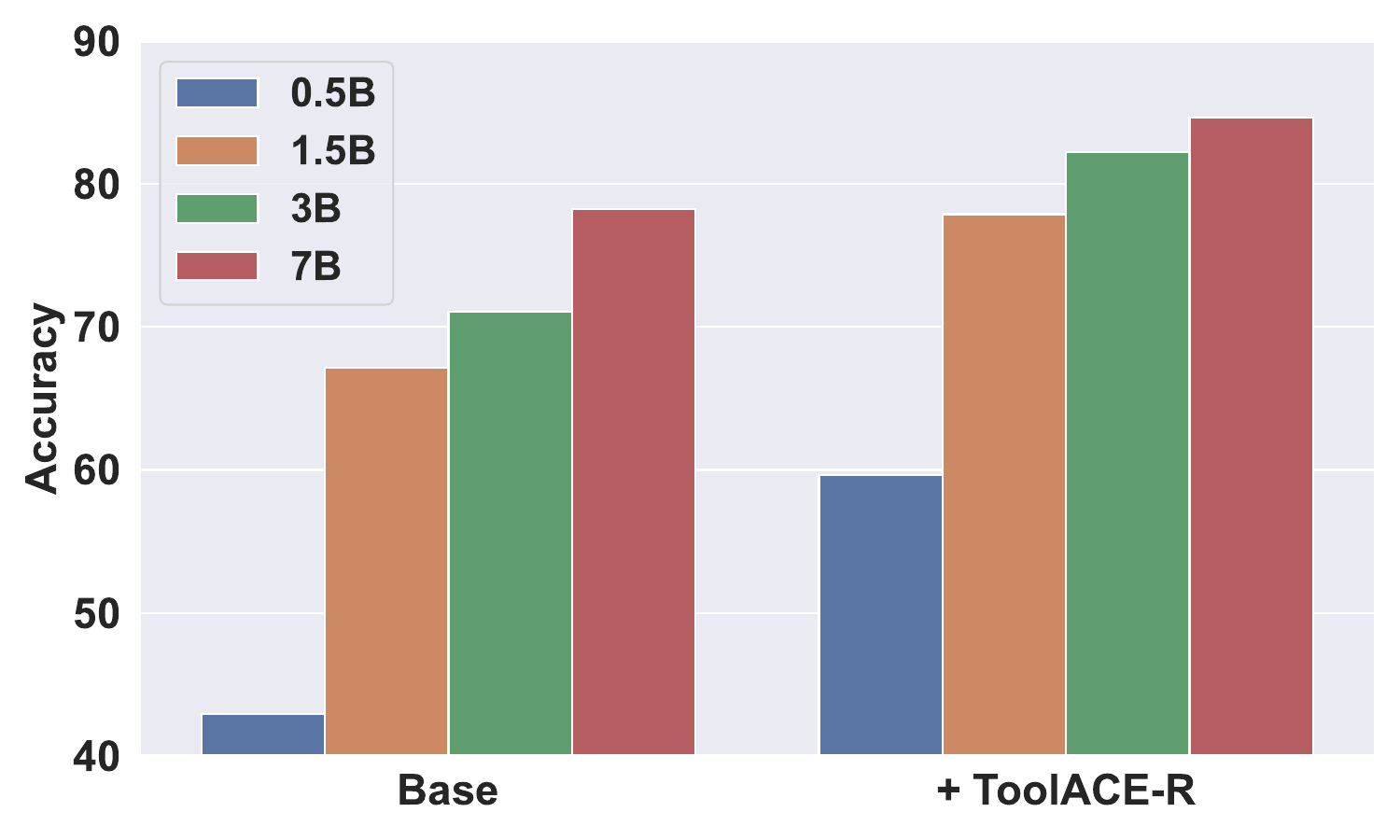}
    \caption{Performance of Qwen2.5-Series. ``Base'' refers to the Qwen2.5-Instuct models without any training, while ``+ \name'' uses our method.}
    \label{fig:scaling}
\end{figure}

\paragraph{Different Backbones.}
To assess the generalizability of \name, we conduct experiments with different base models, including Qwen2.5-7B-Inst and Mistral-7B-Inst-v0.2. The results, along with the original performance and performance after training with the full original training set, are presented in Table~\ref{tab:diff-backbone}. As shown, fine-tuning with the full training set leads to a significant performance boost for both models, particularly for the relatively weaker Mistral-7B-Inst-v0.2, which sees an improvement of over $20\%$. Moreover, our method further enhances performance, using exactly the same original data, demonstrating its effectiveness.

\paragraph{Model Size Scaling.}
We further examine whether our method can be applied to models of varying sizes and whether performance improves as model size increases. We use Qwen2.5-Instruct models of different sizes, including 0.5B, 1.5B, 3B, and 7B, as base models. The results, shown in Figure~\ref{fig:scaling}, display both the original performance and the performance after applying \name. Applying \name consistently improves performance across all sizes, particularly for smaller models like 0.5B and 1.5B. This demonstrates the effectiveness of \name across a range of model sizes.
\section{Conclusion}
In this paper, we introduce \name, a tool learning technique designed to enable LLMs to learn to effectively leverage external tools. \name consists of two components: model-aware iterative training and adaptive self-refine iterative inference. Model-aware iterative training maximizes the model's potential by iteratively selecting the most appropriate samples for training, with a novel definition of sample difficulty. Adaptive self-refinement further enhances performance by allowing the model to autonomously determine when to stop refining its own answers during iterative inference. Experiments demonstrate that \name consistently improves performance across various benchmarks and is effective with different backbone models. Additionally, our analyses highlight the impact of the proposed modules on model performance.

\bibliography{aaai2026}

\section{Algorithm for Adaptive Self-refinement}
Alg.~\ref{alg:inf} describes the process of adaptive self-refine strategy.
\begin{algorithm}
\caption{Adaptive Self-Refine}
\label{alg:inf}
\begin{algorithmic}[1]
\REQUIRE Model $\theta$, query $q$, tool candidates $T$, refine prompt $r$, maximum iteration time $n$
\ENSURE Refined final answer $A$
\STATE $A_0 = f_\theta(\langle q, T \rangle)$ \textcolor{mypink}{\textit{\;// First get the direct answer $A_0$ by directly feeding $q$ and $T$ to the model}}
\FOR{$i = 1$ to $n$} 
    \STATE $A_i = f_\theta(\langle q, T \rangle, A_{i-1}, r)$ \textcolor{mypink}{\textit{\;// Get the refined answer based on the answer of last iteration}}
    \IF{$A_i == A_{i-1}$}
        \RETURN $A_i$ \textcolor{mypink}{\textit{\;// Adative stopping} }
    \ENDIF
\ENDFOR
\RETURN $A_n$ \textcolor{mypink}{\textit{\;// Achieves the maximum iteration time}}
\end{algorithmic}
\end{algorithm}

\section{Detailed Introduction to Benchmarks}
Below we introduce the used benchmarks with more details:

\begin{itemize}[left=0pt]
    \item \textbf{BFCL}~\cite{berkeley-function-calling-leaderboard}: BFCL has progressed over time and is currently at version 4, which includes four main categories: non-live, live, multi-turn, and agent. This paper focuses on the non-live and live categories, which primarily consist of Python-style tool call data. These are further subdivided into four types: Simple, Multiple, Parallel, and Multiple Parallel. 
    \begin{itemize}[left=0.5pt]
        \item \textbf{Single}: A single evaluation is the simplest and most frequently encountered format, where the user provides a single function document, triggering exactly one tool call.
        \item \textbf{Multiple}: The multiple category involves a user query that selects a single tool call from 2 to 4 available function documents. The model must choose the most appropriate tool to call based on the user’s context.
        \item \textbf{Parallel}: A parallel test sample involves triggering multiple tool calls simultaneously in response to a single user query. The model must determine how many calls are needed.
        \item \textbf{Multiple Parallel}: The parallel multiple cases combine the concepts of both parallel and multiple. Here, the model is provided with several function documents, and each corresponding tool call may be invoked zero or more times.
    \end{itemize}

    \item \textbf{ACEBench}~\cite{chen2025acebench}: ACEBench is a comprehensive tool-use benchmark that offers more detailed granularity. It is categorized into three main types: Normal, Special, and Agent. In this work, we focus on the Atom and Single-turn sub-categories within the Normal category. Atom cases involve a set of APIs that contain specific parameter types, such as booleans, enumerations, numbers, lists, and objects. The Single-turn category includes both single and parallel cases.

    \item \textbf{API-Bank}~\cite{li_api-bank_2023}: API-Bank is a dialogue-style tool call dataset, consisting of two settings: Call and Retrieve + Call. In this dataset, the model is tasked with invoking predefined local Python tools based on the user's requirements in the dialogue. Accuracy is assessed by comparing whether the tool's returned values match the ground truth. In this work, we focus on the Call setting in our experiments.

    \item \textbf{ToolAlpaca}~\cite{tang2023toolalpaca}: It introduces a multi-agent simulator that uses GPT-4 to simulate the return values of tools. The model can make modifications and re-call the tool based on the returned values (e.g., error messages). Ultimately, GPT-4 is used to evaluate the accuracy. In our experiments, we does not leverage the returned values for refinement for consistency.
\end{itemize}

\section{Detailed Results in ACEBench}
Table~\ref{tab:acebench-overall} displays the detailed results in each sub-category of ACEBench. As can be seen, in all sub-categories, \name performs better than its base model Llama3.1-8B-Inst. The improvement of Adaptive Self-Refine is not significant.

\section{Results on Multi-Turn Scenarios}
In our main experiments, we conduct experiments on single-turn tool calling for simplicity. We further validate the effectiveness of \name on multi-turn scenarios. We sample 10K multi-turn training instances from ToolACE training data, and apply our model-aware iterative training (including data augmentation with self-refinement, where each turn with tool calling is augmented as a single training instance). We evaluate the performance on Multi-Turn categories of BFCL and ACEBench, as well as $\tau$-Bench~\cite{yao2025tau}, a more challenging benchmark focusing on multi-turn interactions. Table~\ref{tab:mt} displays the results. Clear and consistent gains are shown, demonstrating ToolACE-R’s generalization to longer, complex contexts.

\begin{table}[thb]
\small
\setlength{\tabcolsep}{4.0pt}
\renewcommand{\arraystretch}{1.2}
\centering
\begin{tabular}{@{}lccccc@{}}
\toprule
& \multicolumn{2}{c}{\textbf{BFCL}}          & \multicolumn{2}{c}{\textbf{ACEBench}}              & \multicolumn{1}{c}{\textbf{$\tau$-Bench}} \\ \cmidrule(lr){2-3}\cmidrule(lr){4-5}\cmidrule(lr){6-6}
\textbf{Models} &
\multicolumn{1}{c}{\textit{ST}} &
\multicolumn{1}{c}{\textit{MT}} &
  \multicolumn{1}{c}{\textit{ST}} &
  \multicolumn{1}{c}{\textit{MT}} &
  \multicolumn{1}{c}{\textit{Avg.}} \\ 
  \midrule
\textbf{Llama3.1-8B-Inst }                   	&76.8	&5.6	&50.6	&	26.0&  13.5  \\
\midrule
\textbf{\;\;+ Original Data} &	80.2 &	7.8 &	74.0 &	41.0 & 14.2

 \\
\textbf{\;\;+ \name} &	82.5 &	15.7 &	76.3 &	47.0 & 15.4

 \\
\bottomrule
\end{tabular}
\caption{\label{tab:mt}Multi-turn results on three different benchmarks. ``ST'' is short for ``Single-Turn'', ``MT'' is short for ``Multi-Turn''. ``ST'' for BFCL represents the average of non-live and live categories.}

\end{table}

\begin{table*}[thb]
\small
\setlength{\tabcolsep}{4.0pt}
\renewcommand{\arraystretch}{1.2}
\centering
\begin{tabular}{@{}lccccccccc@{}}
\toprule
& \multicolumn{6}{c}{\textbf{Atom}}          & \multicolumn{2}{c}{\textbf{Single-Turn}}              & \multicolumn{1}{c}{\textbf{Overall}} \\ \cmidrule(lr){2-7}\cmidrule(lr){8-9}\cmidrule(lr){10-10}
\textbf{Models} &
  \multicolumn{1}{c}{\textit{Bool}} &
  \multicolumn{1}{c}{\textit{Enum}} &
  \multicolumn{1}{c}{\textit{Number}} &
  \multicolumn{1}{c}{\textit{List}}&
  \multicolumn{1}{c}
  {\begin{tabular}[c]{@{}c@{}}\textit{Object}\\ \textit{Short}\end{tabular}}&
  \multicolumn{1}{c}{\begin{tabular}[c]{@{}c@{}}\textit{Object}\\ \textit{Deep}\end{tabular}} &
  \multicolumn{1}{c}{\textit{Single}} &
  \multicolumn{1}{c}{\textit{Parallel}} &
  \multicolumn{1}{c}{\textit{Overall}} \\ 
  \midrule
\textbf{GPT-4o-2024-11-20 }                   & 94.00	& 94.00	&98.00	&94.00	&64.00	&96.00	&84.00	&	72.00&  87.00  \\
\textbf{GPT-4o-mini-2024-07-18   }          	&88.00	&94.00	&98.00	&90.00	&44.00	&92.00	&77.00	&70.00	& 81.63 \\

\textbf{Llama3.1-8B-Inst}  &50.00	&74.00	&48.00	&60.00	&	30.00&44.00	&55.00	&44.00	& 	50.63
 \\
\midrule
\textbf{\name} &	88.00 &	96.00 &	94.00 &	94.00 &	50.00 &	92.00 &	83.00 &	74.00 & 83.88

 \\
 $\;\;$\textbf{+ Adaptive Self-Refine} &	88.00 &	96.00 &	94.00 &	94.00 &	50.00	&	94.00 &	84.00 &	72.00 & 84.00

 \\
\bottomrule
\end{tabular}
\caption{\label{tab:acebench-overall}Detailed accuracy comparison on ACEBench. }

\end{table*}

\begin{figure*}[!h]
  \begin{tcolorbox}[colback=blue!5!white,colframe=blue!75!black,fontupper=\small,fontlower=\small]
    You are an expert in composing functions. You are given a question and a set of possible functions. Based on the question, you will need to make one or more function/tool calls to achieve the purpose. If none of the function can be used, point it out. If the given question lacks the parameters required by the function, also point it out. You should only return the tool call in tools call sections. \\\\
    If you decide to invoke any of the function(s), you MUST put it in the format of: \\  $[$func\_name1(params\_name1=value1, params\_name2=value2,...), func\_name2(params)$]$ \\\\
    
    You should not include any other text in the response.
    Here is a list of functions in JSON format that you can invoke:\\
    \{candidate tools\}\\
    \\
    \{other information\}
  \end{tcolorbox}
  \caption{The system prompt for tool calling. ``\{candidate tools\}'' is replaced with actual tool descriptions. ``\{other information\}'' is replaced with some other information in the case, e.g., date. If no other information, it is null.}
  \label{fig:system-prompt}
\end{figure*}

\begin{figure*}[!h]
  \begin{tcolorbox}[colback=blue!5!white,colframe=blue!75!black,fontupper=\small,fontlower=\small]
    Please refine your answer. Directly output the refined answer, or the original answer if you think it is already perfect.
  \end{tcolorbox}
  \caption{The refine prompt for tool calling.}
  \label{fig:refine-prompt}
\end{figure*}

\section{Limitations}
We summarize the limitations of this work in the following two points:

First, due to resource constraints, we conduct experiments only on models with up to 8B parameters, using LORA, without applying our \name method to larger LLMs. However, as shown in the scaling performance results in Figure~4 of main content, we have not yet observed any bottlenecks preventing larger models from benefiting from our \name method.

Second, we focus exclusively on single-turn tool-calling scenarios, to maintain simplicity and stability. We argue that single-turn interactions are a fundamental component of tool learning. The substantial improvements observed across several representative benchmarks highlight the effectiveness of our approach. We also supplement our evaluation with multi-turn scenarios (Table~\ref{tab:mt}), which provides additional evidence of its effectiveness.

\section{Broader Impacts}
The development of \name represents a step forward in enabling more autonomous and capable large language models (LLMs) through efficient and scalable tool learning. By reducing reliance on externally curated feedback and training with LORA in small LLMs, \name lowers the barriers to deploying high-performing LLMs in resource-constrained environments. 

However, the growing autonomy of LLMs in generating and refining tool-use behavior also raises potential concerns. As models become better at self-directed decision-making, ensuring safety, accountability, and alignment with human values becomes increasingly important. Misuse of such systems, e.g., in automating high-stakes tasks without sufficient oversight, could lead to unintended consequences.

Moreover, as LLMs become more proficient at tool interaction, there is a risk of amplifying biases or inaccuracies embedded in external tools or data sources. While we focus on technical robustness and efficiency, further research should explore mechanisms for value alignment, transparency, and fail-safe design in autonomous tool-using agents.

\section{Prompts and Data Example}
We show the detailed system prompt and refine prompt in Figure~\ref{fig:system-prompt} and Figure~\ref{fig:refine-prompt}, respectively.
Note that the displayed prompts are those used during inference. For the training data, we diversify the system prompts to ensure better generalizability.

Figure~\ref{fig:data-example} is an example of our refine data. The model will only learn the last turn, with previous turns as context.

\begin{figure*}[!h]
\begin{tcolorbox}[colback=blue!5!white,colframe=blue!75!black,fontupper=\small,fontlower=\small]
\textbf{[System]} \\
You are an expert in composing functions. You are given a question and a set of possible functions. Based on the question, you will need to make one or more function/tool calls to achieve the purpose. If none of the function can be used, point it out. If the given question lacks the parameters required by the function, also point it out. You should only return the tool call in tools call sections. \\\\
If you decide to invoke any of the function(s), you MUST put it in the format of: \\ $[$func\_name1(params\_name1=value1, params\_name2=value2,...), func\_name2(params)$]$ \\\\

You should not include any other text in the response.
Here is a list of functions in JSON format that you can invoke:
\begin{verbatim}
{
    "name": "bookFlight",
    "description": "Book a flight for a specified destination",
    "parameters": {
        "type": "dict",
        "properties": {
            "origin": {
                "type": "string",
                "description": "The departure airport or city"
            },
            "destination": {
                "type": "string",
                "description": "The destination airport or city"
            },
            "departure_date": {
                "type": "string",
                "description": "The date of departure (in YYYY-MM-DD)"
            }
        },
        "required": [
            "origin",
            "destination",
            "departure_date"
        ]
    }
}
\end{verbatim}
\begin{verbatim}
...

\end{verbatim}
The current time is 2024-10-02 18:18:11.\\

\textbf{[User]} \\
I need to book a flight from New York to London for a business trip tomorrow.\\

\textbf{[Assistant]} \\
$[$bookFlight(origin="New York", destination="London", departure\_date="2024-10-03")$]$\\

\textbf{[User]} \\
Please refine your answer. Directly output the refined answer, or the original answer if you think it is already perfect. \\

\textbf{[Assistant]} \\
$[$bookFlight(origin="New York", destination="London", departure\_date="2024-10-03")$]$
\end{tcolorbox}
\caption{One example for refine data. In this example, the assistant does not change the tool invocation as it is correct already. Through this kind of data, the model is able to learn when to stop iteration during inference.}
\label{fig:data-example}
\end{figure*}

\end{document}